\documentclass[a4paper, 10pt, conference]{ieeeconf} 
\AtBeginDocument{\let\autocite\cite}
\let\tightlist\relax

\IEEEoverridecommandlockouts                              % This command is only needed if                                                           % you want to use the \thanks command

\usepackage{balance}
\usepackage{url}
\usepackage{cite}
\usepackage{graphicx} % for pdf, bitmapped graphics files
\usepackage{flushend}
\usepackage{booktabs}
\usepackage{tabularx}
\usepackage{tikz}
\newcommand\copyrighttext{%
  \footnotesize \textcopyright 2024 IEEE. Personal use of this material is permitted.
  Permission from IEEE must be obtained for all other uses, in any current or future
  media, including reprinting/republishing this material for advertising or promotional
  purposes, creating new collective works, for resale or redistribution to servers or
  lists, or reuse of any copyrighted component of this work in other works.}
\newcommand\copyrightnotice{%
\begin{tikzpicture}[remember picture,overlay]
\node[anchor=south,yshift=10pt] at (current page.south) {\fbox{\parbox{\dimexpr\textwidth-\fboxsep-\fboxrule\relax}{\copyrighttext}}};
\end{tikzpicture}%
}
\begin{document}
\IEEEoverridecommandlockouts
\overrideIEEEmargins

\title{\LARGE \bf
  Toward Anxiety-Reducing Pocket Robots for Children
}

\author{Morten Roed Frederiksen,$^{1}$ Kasper Stoy,$^{2}$ & Maja Matari\'{c}$^{3}$
  \thanks{$^{2}$Kasper Stoy {\tt\small ksty@itu.dk} is affiliated with the REAL lab at the Computer science department of The IT-University of Copenhagen, Rued Langgaards vej 7, 2300 Copenhagen S. $^{1}$Morten Roed Frederiksen {\tt\small mrof@itu.dk} and $^{3}$Maja Matari\'{c} are affiliated with the Interaction Lab in the Computer Science Department at the University of Southern California.}
}

\maketitle

\copyrightnotice
\begin{abstract}
A common denominator for most therapy treatments for children who suffer from an anxiety disorder is daily practice routines to learn techniques needed to overcome anxiety. However, applying those techniques while experiencing anxiety can be highly challenging. This paper presents the design, implementation, and pilot study of a tactile hand-held pocket robot “AffectaPocket”, designed to work alongside therapy as a focus object to facilitate coping during an anxiety attack. The robot does not require daily practice to be used, has a small form factor, and has been designed for children 7 to 12 years old. The pocket robot works by sensing when it is being held and attempts to shift the child's focus by presenting them with a simple three-note rhythm-matching game. We conducted a pilot study of the pocket robot involving four children aged 7 to 10 years, and then a main study with 18 children aged 6 to 8 years; neither study involved children with anxiety. Both studies aimed to assess the reliability of the robot's sensor configuration, its design, and the effectiveness of the user tutorial. The results indicate that the morphology and sensor setup performed adequately and the tutorial process enabled the children to use the robot with little practice. This work demonstrates that the presented pocket robot could represent a step toward developing low-cost accessible technologies to help children suffering from anxiety disorders. 
\end{abstract}

\section{Introduction}

Anxiety disorders cause children as young as 6 years old to avoid
activities of daily living, including getting out of bed,
eating, and going to school, thereby disrupting their lives, and the
lives of their families
\autocite{Schneider2012SeparationAD,Jarrett2015GeneralizedAD,Norris2016UpdateOE,StuartParrigon2016FamilyPI}.
Data from a 2019 U.S. study showed that depression and anxiety disorders
were at the top of the list of disabling mental illnesses and that more
than 1 out of 11 children between 3-17 years old were diagnosed with
some form of anxiety disorder
\autocite{VosLimAbbafatiAbbasAbbasiAbbasifardAbbasiKa20,Bitsko2022MentalHS}.
Cognitive behavior therapy (CBT) is a widely used treatment for
anxiety disorders across this age span \autocite{Whiteside2019AMT}. CBT
coping strategies include changing the focus of attention, such as using
distraction imagery associated with something positive, listening to
music, and repeating a mantra with a positive association
\autocite{avants1990,lvarezPrez2022EffectivenessOM,LYNCH2018101}. To be
effective, CBT techniques often require daily practice
\autocite{Lorimer2020PredictorsOR,Tyrer2006GeneralisedAD}. Recent robotics research has investigated helping children suffering from anxiety disorder by introducing socially assistive robots as training companions
that train the children to navigate in various anxiety-inducing contexts
\autocite{Zhu2020EffectOS}. Such approaches have the potential to increase children's
adherence to daily training. This paper
presents the design, implementation, and initial study of
``AffectaPocket'', a pocket robot created with the goal of helping children to manage an anxiety attack in the moment, aimed at being useful to children who have not had the ability to practice or master CBT or other coping strategies. 

\label{Technical Approach}
\begin{figure}[h] \centering \includegraphics[width=0.492\textwidth]{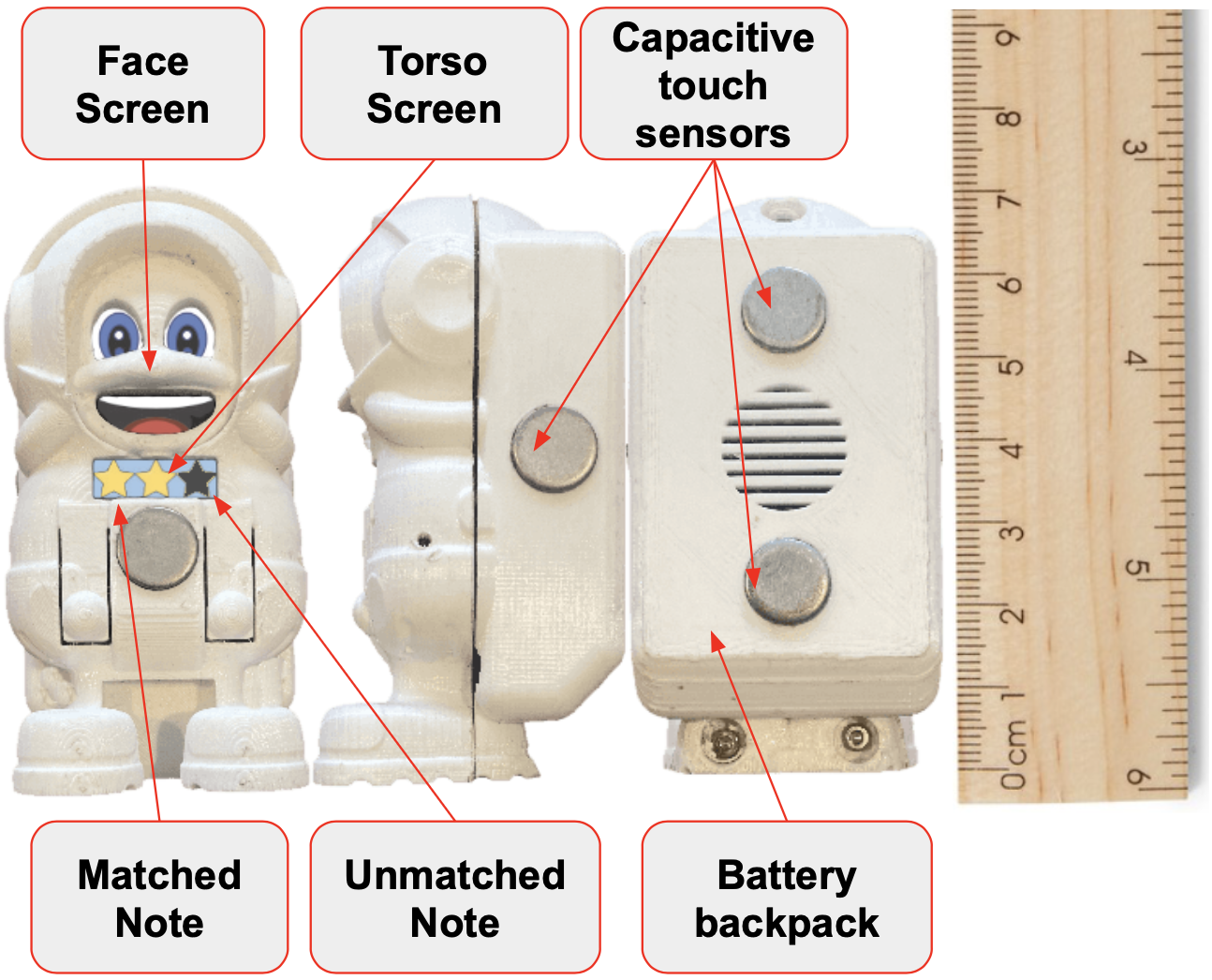} \caption{The AffectaPocket robot (shown with a ruler for scale). The interaction with the robot is tactile and private, and can be conducted in a child's pocket. The robot is equipped with five touch-capacitive sensor plates (1 at the front, 1 at each side, 2 at the back) and driven by an ESP32 development board. When grasped, the robot vibrates in a rhythmic pattern. The child then tries to match the pattern by grasping and releasing the robot, thereby focusing on the interaction and away from the source of anxiety. The robot's face screen displays black stars for each rhythm note and turns them into gold stars as the child matches each note.  This visual interface is useful for training, until the child is confident enough to rely on tactile vibration cues alone.}
\label{affecta_pocket}
\end{figure}
The robot (shown in Figure \ref{affecta_pocket}) is designed for
children in the target age range of 7 to 12, and is a result of a
participatory co-design that involved end-users in the process of technology design
\autocite{burkett2012introduction,lee2008design,jessen2018creating}.The
design process involved obtaining feedback from children in early
evaluations and conducting an initial pilot study of the robot's design and
functionality. The lead author's and friend's children,
aged 7 to 11, and spanning different genders, were engaged in the co-design.
This paper outlines the early results from the initial tests of the pocket robot.

AffectaPocket aims to offer an immediate diversion of attention in stressful situations. As a toy-like robot, it is designed to motivate the child to
interact with it. The robot senses when it is being grasped or held and
responds in a way that aims to draw the child's attention away from the
source of stress and toward itself, in a private way, from the pocket,
not in plain view of other children and adults \autocite{avants1990}.
When grasped, the robot initiates a simple rhythm-matching game that
requires the child to concentrate on repeating a randomly generated
3-note vibrating rhythm the robot silently generates. The child attempts
to match the rhythm by feeling the tactile vibrations of the robot and
grasping and releasing the robot in a similar temporal pattern.

The study presented in this paper evaluated the pocket robot's tactile
features, touch sensor input, and initial user impressions of its morphology. Together, these features outline the most important
affordances of the robot that encompass how it is used.

\section{Background on anxiety}
%Our results indicate that the AffectaPocket has a morphology that allows children to comfortably keep it in their pockets for a full day and that the sensors allow it to consistently register when children grab it to initiate the distraction features with a limited amount of false positive sensor events. The results also indicate that the tutorial process created to convey the rules of the rhythm-matching game works as intended as the children transitioned into using the device concealed in their pockets within a few attempts requiring no previous experience with tactile devices. The overall takeaway gained from the experiments is that several areas of the robot could be improved but AffectaPocket could be a step in the right direction toward providing additional aid in a dire situation for many children to whom there is little else to rely on but therapy or medication.

A global study from 2019 showed that depression and anxiety disorders
reside at the top of the list of disabling mental illnesses
\autocite{VosLimAbbafatiAbbasAbbasiAbbasifardAbbasiKa20}. Research shows that causes are multifold, making it difficult to introduce prevention measures \autocite{Velting2001CurrentTI}. There is also a growing
concern over the impact of social media use on developing
anxiety disorders in both children and adolescents
\autocite{Kowalchuk2022AnxietyDI}. Many of those
who could benefit from treatment are not diagnosed
\autocite{Baldwin2010PharmacologicalTO}.

%\subsection{How it manifests}

The symptoms of anxiety disorder are varied. Generalized anxiety disorder
manifests as anxiety unrelated to encountered stressful scenarios but
can be increased by events and may make children feel
threatened, irritable, and unable to sleep. It may also manifest with
physical symptoms such as heart pounding, sweating, and muscle tension
\autocite{Tyrer2006GeneralisedAD}. Anxiety is difficult to measure and
evaluate even with self-reporting measures such as the anxiousness scale
\autocite{Leary1983SocialAT,Leary1993TheIA}. Besides CBT and related therapies, methods utilizing biofeedback and
meditation have been shown to increase the impact of therapies
\autocite{Ratanasiripong2015StressAA}.  Recent research has
investigated finding Biomarkers for Generalized anxiety disorder (GAD), focusing on blood tests and neuroimaging to discover biologically-grounded treatments
\autocite{Maron2017BiologicalMO}.  

Although not directly tested on a vulnerable population to date, the design choices in this work were informed by the specific needs of those experiencing anxiety. This research presents a novel technological approach toward assisting young children with in-the-moment anxiety management.

%\subsection{A new approach}

\section{System overview}

\label{system_overview} 
\begin{figure}[h] \centering \includegraphics[width=0.492\textwidth]{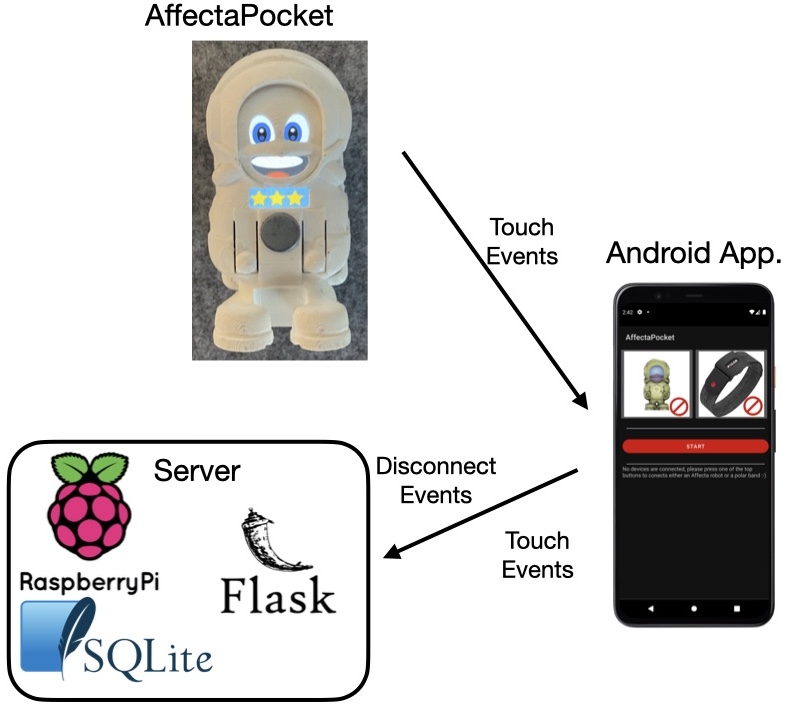} \caption{The three-tier system setup used to gather data for the experiments. The Android application (right) discovers and connects to AffectaPocket. The pocket robot (top) notifies the application of all touch and release events. The application connects to a REST server running on a Raspberry Pi 4 (left). The Android application sends all touch events to the server. The server stores all values in an SQlite database.}
\label{three_tier}
\end{figure}

AffectaPocket is designed to function as an autonomous entity. The pocket robot employs capacitive touch sensors for environmental interaction and integrates a rudimentary controller coupled with a resonant actuator to facilitate user communication. With sensors, control, and an actuator, AffectaPocket contains the core elements of a simple robot. To facilitate data collection, the robot is tethered to an Android companion application establishing communication through a Python Flask server running on a Raspberry Pi 4 system on a chip (SOC) board in a three-tier system architecture. \autocite{Relan2019DeployingFA}. The server stores all received sensors values in an SQLite database, with a timestamp for each event. The system overview can be
seen in Figure \ref{three_tier}.

While the Android application and server functionality were vital for
testing the design of the device, the intention is for the pocket robot to work as a
self-contained device in future versions. This section focuses
on the core functionalities and design choices for the pocket robot.

\subsection{Morphology of the pocket robot}

In the development of AffectaPocket, consideration was given to its suitability for children with anxiety disorders who may prefer to avoid drawing attention to themselves. Although the robot is minimalistic, primarily consisting of sensors, controller, and a vibration-inducing actuator, the choice to label this device a robot rather than a therapeutic tool is deliberate and aimed at enhancing engagement by invoking a sense of personality and agency typically associated with anthropomorphic entities. The use of anthropomorphic language not only invites children to interact with the robot more naturally but also encourages them to attribute human-like qualities to the robot. This approach aims to foster an anthropomorphic interpretation that could positively shape the interactions with the robot. By framing the robot in these characteristics when describing and introducing it, we intend to prime the children to view the device as a robot and potentially as a social partner, with the goal of enriching their learning experience and interaction quality \autocite{8172452, darling2015s, riek2009anthropomorphism, onnasch2021taxonomy}. 
The operational principle of the robot centers on tactile interaction. The robot detects when a child touches or grasps it, acting as an intervention tool during anxiety episodes by using an attention diversion mechanism consisting of a rhythm-matching game that requires the child to focus on repeating a randomly generated 3-note vibrating
rhythm. To do so, the child must recognize the rhythm by feeling the
tactile vibrations of the grasped robot and then grasp and release the robot
in a similar pattern.  The robot makes no sound; because of the nature of anxiety, we believe that the technology developed to help coping should aim to be discrete. 

AffectaPocket robot is a 7cm x 4cm x 3cm 3D-printed device depicting a robot with a backpack. The 3D-printing material is PLA; the outer shell is assembled by two
individual halves. The back piece has a square backpack with room
to house two standard AAA batteries. The front
piece has holes for the eyes and mouth and a square hole for a screen
on the robot's torso. The front piece houses the main SOC board
which consists of an ESP32 development board with an attached LCD screen
with a 272x180 pixel resolution. The board has Bluetooth low energy
connectivity and registers as a peripheral with a single connectable
service that can notify any connected devices of changes to the robot's sensor
values.

\subsubsection*{Participatory design evaluation}
In the development of the pocket-sized robot, three children were instrumental by actively participating in the design process. The project employed the participatory design approach, specifically the cooperative design method, which involves potential users in the creation process to ensure the final design meets their needs and preferences \autocite{Velden2014}. This method is particularly effective for fostering a collaborative environment and ensuring user-centered design outcomes \autocite{Mackay2020, spinuzzi2005}. The objectives of this project were threefold: first, to optimize the robot's dimensions and form for a smooth fit within a child's pocket; second, to equip the robot with sufficient touch sensor coverage to ensure that interactions with children of the targeted age group were detected by at least one sensor as they grasped the robot; and third, to design a robot form that would spark the imagination of children and encourage interaction.

To achieve the third objective, the children participated in brainstorming sessions to envision several imaginative robot forms. The final design emerged as a hybrid of various animals—a tortoise and a bird—integrated into a humanoid astronaut suit. The astronaut suit features were inspired by a pre-fabricated 3D-printing test figure that will be changed to a different design in future iterations \autocite{matterhackers}. The overall design of the robot was intended to invite playful interactions, aligning with the notion that the design should encourage play activities. The backpack was included to reflect an astronauts suit, but also for housing the batteries forpowering the robot. This iterative design process involved multiple rounds of design, development, and testing, during which the children carried the robot in their pockets during 1-hour play sessions. This cycle was repeated three times to refine the robot's design, using the participatory design method to maintain a focus on user-centered development.

\subsubsection*{Learning of the Rhythm-Matching Game through Visual Cues}

Central to the robot's design is the main screen located on its torso, whose role is to aid in instructing children how to play the rhythm-matching game. The screen visually represents the game's rhythms by animating stars that align with each vibrational note, thereby aiding in pattern recognition and enhancing the children's understanding of the game mechanics. The visual representation changes dynamically as each note in the rhythm is matched, allowing children to quickly grasp the game's concept. The rhythm-matching game, described in Section \ref{rythm_matching}, is designed to serve as an attention-diverting activity for children during anxiety episodes. Successful replication of the robot's three-note sequence by the child triggers the generation of a new rhythmic pattern for the child to match.
This interactive game is intended to engage the child's attention and concentration on the rhythmic patterns without necessitating regular practice.

\subsubsection*{The Robot's Facial Expressions}

AffectaPocket is programmed to exhibit facial expressions, a feature that fosters anthropomorphic perceptions by its users \autocite{song2020trust}. This aspect of the design provides a straightforward method for conveying the robot's status, aligning with children's innate tendency to interpret and respond to facial expressions during the development of early social-emotional skills \autocite{field1977effects,tronick1989infant}. 
While these animated expressions do not directly contribute to the robot's primary function of alleviating anxiety, they play a significant role in animating the otherwise inanimate 3D-printed form, thereby enhancing user engagement. The rationale behind incorporating facial animations into the robot's design is to promote a fun and engaging interaction that will stimulate children to have an interest in using and
adopting the technology.
Additionally, the idea behind the robot's form is that children may feel more inclined to use a robot that
they see as a small pocket-sized friend rather than a non-anthropomorphic
3D-printed robot, since research has shown that both anthropomorphic language about robots \autocite{8172452} and anthropomorphic robot features can serve to improve trust and likeability of robots in some contexts \autocite{Barone2023CallIR, Schreiter2022TheEO}. Future iterations of the pocket robot might incorporate adaptive mood and expression changes in response to the child's progress in managing anxiety. This would reinforce the child's perception of the robot as a friend that goes through difficult situations and grows stronger alongside them.  Therefore, facial expressions are useful in the initial acceptance and engagement with the pocket robot, even if its typical use is concealed in the pocket.

\subsection{Tactile communication}

The robot has a small vibration motor mounted at the back of the ESP32
board that conveys the notes of the rhythm game and
informs the child when they have successfully matched a rhythmic
pattern. The
current vibration motor only works with binary patterns to display
either on or off and cannot convey shorter rhythm patterns below 80ms, limiting the complexity of the pattern that can be conveyed to the child.
The rationale for the rhythm-matching game's design is to make it both enjoyable and challenging, motivating children to stay engaged and complete each pattern.

\subsection{Touch sensors}

There are five capacitive touch sensor surfaces on the robot, two at the back, one at each side, and one at the front below the torso screen (as shown in Figure \ref{affecta_pocket}). These are
10mm diameter metal surfaces that register when the user grasps or touches
the robot and provide a measured capacitance value between 0 and 100. The robot checks for changes in capacity with a refresh rate
of 10hz. When the value falls below a preset threshold value, the
robot registers the surface plate as touched. Each touch plate has three
states: Touched, Released, and Not Changed. With these states, each touch
surface can measure when a touch was initiated, and when it was released,
and calculate the total length of the touch. A challenge in using
capacitive touch sensors is that compared to mechanical binary switches,
they may accidentally provide false positive touch events if not
calibrated correctly to the specific task or working environment. However, compared to mechanical switches, capacitive touch sensors offer a more intuitive interaction for children, as they require less physical engagement, potentially enhancing usability in our robot design.

\section{Pilot Study}

As a first step toward an evaluation of the design with children with anxiety, we evaluated the system with typically developing children. 
We performed a pilot study with three participants followed by a main study with 18 participants, all aged 7 to 11. Both the pilot and the main study
were designed to provide initial indications of the functionality of the
device and to reveal any practical design flaws of AffectaPocket. The studies focused on verifying two aspects of the design:

\begin{itemize}
\tightlist
\item
  The physical shape and capacitive touch capabilities of the robot
  during extended use;
\item
  The tactile and vibrational capabilities of the robot and the ease of use for children.
\end{itemize}

We investigated the first aspect by monitoring children as
they used the device in play sessions of various lengths and with
varying intensities of physical activity. The second aspect was tested
monitoring children as they completed a short tutorial process and measuring
their skill progression.

\subsection{Physical shape and capacitive touch capabilities}

As noted earlier, making sure AffectaPocket is discrete was an important design constraint, warranting
a morphology with limited physical proportions that enabled the robot to fit
inside pockets of children aged between 7 to 12 years. The physical
shape of the device had to be comfortable enough to carry 
inside a pocket for extended periods. Additionally, the capacitive touch plates on
the device had to be designed to not trigger false positive touch
events in a pocket close to the skin, and
when children were physically active.

To investigate this, AffectaPocket was carried in the trouser pocket on
separate occasions by two 7-year old children. The device and the
children were monitored through the following sessions:

\begin{enumerate}
\def\labelenumi{\arabic{enumi}.}
\tightlist
\item
  4 x 2-hour play sessions;
\item
  2 x 8-hour full-day sessions.
\end{enumerate}

In each of the play sessions, a child carried AffectaPocket in their pocket. The child was informed that the robot needed
their help and it would make it happy if it could spend a few hours
inside their pocket. At each start of
the event and each half hour of the first event, the children were
asked to grasp the robot and let it go, to ensure that the robot
could robustly detect those events. The play sessions were conducted in
the homes of the participating children. No further instructions
given to the children nor was their movement restricted. 

The outside play
sessions were recorded near the children's homes and consisted of
physical activities such as ball games and laser tag tournaments.
After each session, the children were asked if they had grasped the robot during the session to filter out false positives. 

The full-day
events were conducted from 8 am to 4 pm. Each child was informed that
the robot would like to spend a full day in their pocket. No further
instructions were given to the child and they were not asked to grasp the
device at any point. The mobile phone tethered to the robot was
strategically placed at a central location to ensure the children 
remained within the signal limit of Bluetooth LE.

\subsection{The tactile and vibrational capabilities and ease of use}

\label{rythm_matching} Helping children understand how to use the device
properly is vital for its effectiveness in helping to cope with an
anxiety disorder. In the pilot study, we evaluated how the device
conveyed the rhythm-matching game to children through a 10
minute tutorial. The tutorial process was evaluated by 4 different
children aged 7 to 10.\\

\noindent{\bf Tutorial steps}

\begin{enumerate}
\item \noindent{The tutorial process was initiated when a child grasped AffectaPocket.}
\item \noindent{The robot generated a random three-note rhythm through vibrations.}
\item \noindent{The center screen on AffectaPocket displayed black stars in a similar rhythmic pattern.}
\item \noindent{As the child grasped the robot, the grasp timing was compared to the generated pattern.  }
\item \noindent{When a note was matched, a black star was replaced by a gold star on the robot's screen. }
\item \noindent{When all three notes were matched, the face screen displayed a smiling face and the robot vibrated continuously for three seconds, then stopped. }
\item \noindent{The process repeated from the beginning and continued until the child let go of the robot for 8 seconds.}
\end{enumerate}

Three different phases were completed by each child, first using the
visual cues from the center screen, then using no visual cues (the
children were blindfolded), and then using just contact from a pocket.
Each phase continued until the child matched at least three generated
rhythmic patterns and would usually last between 1-2 minutes. A single note was considered successfully matched if the child's press was within 40\% of the generated note length.
The number of attempts needed to complete the match was recorded, as was
the average precision on the three matched notes.

The pilot study showed that the children comprehended the tactile
functionality offered by the robot, as shown in Figure \ref{pattern_match},
measured by the progress made in each phase of the tutorial process. 
As the game's complexity escalated through its various phases – starting with visual cues, followed by playing blindfolded, and finally with the robot in the pocket – the number of attempts required to complete each game within each phase decreased, as seen in Figure \ref{pattern_match}. This trend indicates a progressive improvement in the childrens' skill levels as they advanced through each stage of the tutorial.

\section{Pilot study results}

Our pilot study evaluation provided the following findings and insights.

\subsection{The physical shape and capacitive touch capabilities}

With the initial play session experiments, we wanted to investigate if
the robot would register any touch events from the capacitive touch
sensors outside of the instructed half-hour grasp events planned as a
part of the experiment. The initial experiment containing the four
two-hour play sessions revealed that the planned grasp of the device at
the start and each half hour of the experiment were robustly detected, as
shown in the top part of Table \ref{first_test}, with a total of 19
events registered at the initial instructed grasp, 11 at the second, 6 at
the third, and 6 at the last. There were two false positives detected
during the play sessions.

\begin{table}[]
\begin{center}
\begin{tabular}{@{}ll@{}}
\toprule
\textbf{Play session 4x2h.} & \textbf{Avg. touch events per session} \\ \midrule
1st grasp event & 9,5 \\
2nd grasp event & 5,5 \\
3rd grasp event & 3 \\
4th grasp event & 3 \\ \midrule
Avg. reconnects per session & 4 \\
Total errors & 2 \\
\bottomrule
\end{tabular}
\end{center}
\caption{The logged touch events from the initial four 2-hour play sessions. The grasp events were introduced each half hour and the children were asked to grasp the device in their pocket and let go again.}
\label{first_test}
\end{table}

The second part of the pilot experiment ran for 8 hours each, on two separate days. The participants were not confined to a specific room or location and did not receive further instructions during the experiment. As Table \ref{second_test} shows, a total of 16 touch events were registered. In
both recorded runs, the timestamp of the registered touch reveals that
these events happened within the initial twenty seconds of the test (8
in the first, 10 in the second).

\subsection{Tactile capabilities and ease of use}

\begin{figure}[h] \centering \includegraphics[width=0.49\textwidth]{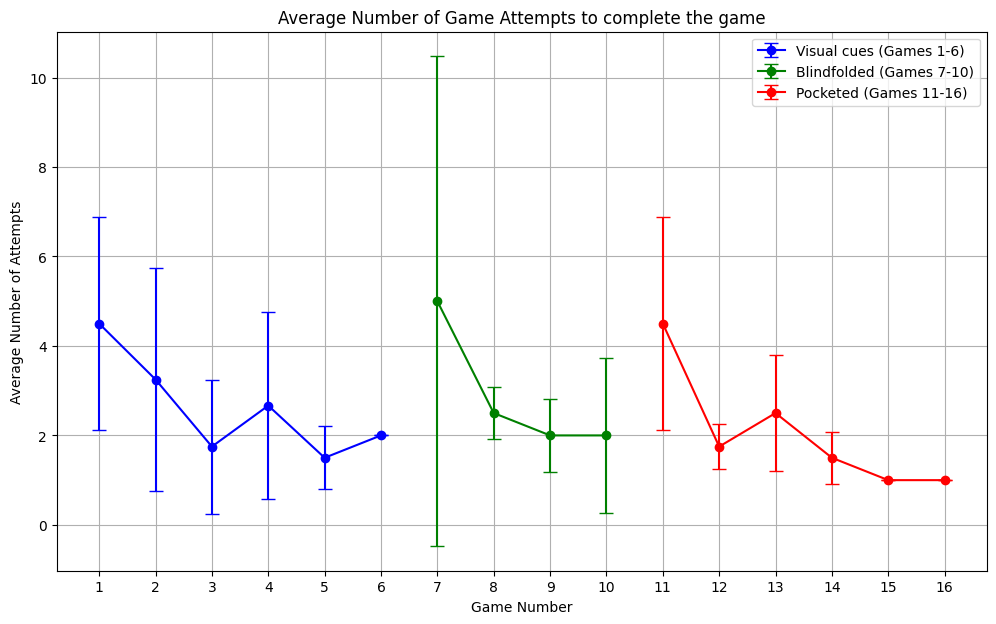} \caption{Pilot Study: The average number of attempts it took the children to complete each presented rhythmic pattern. The x-axis shows the presented pattern number. The y-axis shows the average number of attempts needed to successfully match the pattern. Each series represents a phase in the training process, starting with the initial training phase (blue), the blindfolded phase (green), and the pocket phase (red).}
\label{pattern_match}
\end{figure}

The pilot experiment also investigated the challenges in  learning to use the AffectaPocket and potential
challenges in conveying the rhythmic pattern-matching functionality to children. The experiments entailed running a tutorial process with
three separate phases with four participating children. Three of the
children were 7 years old (two male, one female) and one child was 11 years old (male). Prior to the study, informed consent was obtained from the parents and assent from the children. The results can be seen in Figure \ref{pattern_match}; the figure indicates
progress made by the children as they increased their skill level in
matching a generated rhythmic pattern. Each phase presents a higher
difficulty level: the initial phase presented a visual cue, the second phase had to be completed
blindfolded (no visual cues), and the third phase involved using the device in
their pockets. Although the difficulty increased with each phase, as shown in Figure
\ref{pattern_match}, the children demonstrated similar skill level progress measured as the
number of attempts needed to complete each presented
pattern. At the onset of each new phase and therefore also a higher level of difficulty, an increase in the number of attempts to complete the game was observed. This pattern suggests that simplifying the tutorial by reducing the number of phases could be beneficial in the extended study. 
The precision percentage indicates how well the
children matched each note. The average precision percentage remained stable at an average of 21,99\% (stdev:4.54)
in the tutorial phase, 19,84\% (stdev: 4.46) in the blindfolded phase,
and 19,97\% (stdev: 4.47) in the pocket phase.
\begin{table}[]
\begin{center}
\begin{tabular}{ll}
\toprule
\textbf{Number of hours in session} & 8hrs.\\ \midrule
Total touch events & 16 \\
\bottomrule
\end{tabular}
\end{center}
\caption{The logged touch events from the two 8-hour experiments. }
\label{second_test}
\end{table}

\section{Main Study}

Following the promising results of the pilot study, a main study was conducted to further evaluate the tactile capabilities of AffectaPocket. This study involved 18 children recruited from a first-grade class at a local school. Prior to the study, informed consent was obtained from the parents and assent from the children.

\subsection{Background and Study Adjustments}

The pilot study revealed that while children could effectively learn the tactile game, they faced challenges in regaining their rhythm-matching skills when transitioning from visual cues to blindfolded play, and finally to using the robot in their pockets. In light of these findings, we opted to omit a phase (the blindfolded) from the tutorial, aiming to explore the effectiveness of the tactile game with one less transition. Moreover, considering the impracticality of blindfolding in non-laboratory settings, this adjustment would also potentially enhance the overall usability of our approach. The main study was designed with a modified tutorial process as follows:

\begin{itemize}
    \item Each child was first given one attempt to match the rhythm by following the visual cues displayed on the robot.
    \item Immediately after the visual cue attempt, the child was asked to play the game with the robot held either behind their back or concealed in their pocket. This was needed to observe the children's adaptability to tactile feedback without prior extensive practice.
    \item  The children played the game for a duration of two minutes. During this time, two key metrics were recorded: the number of completed games, and the number of attempts taken to complete each game.
\end{itemize}

Despite the shortened tutorial process, we anticipated that children's learning with AffectaPocket would mirror the pilot study, with an initial high number of attempts to complete the first game due to unfamiliarity with tactile feedback and, as children progressed, a decrease in subsequent games. Additionally, we predicted that varied developmental skill levels among the children would result in initial differences in performance, but those would diminish over time. 
\section{Results of the Main Study}

\begin{figure}[ht] \centering \includegraphics[width=0.49\textwidth]{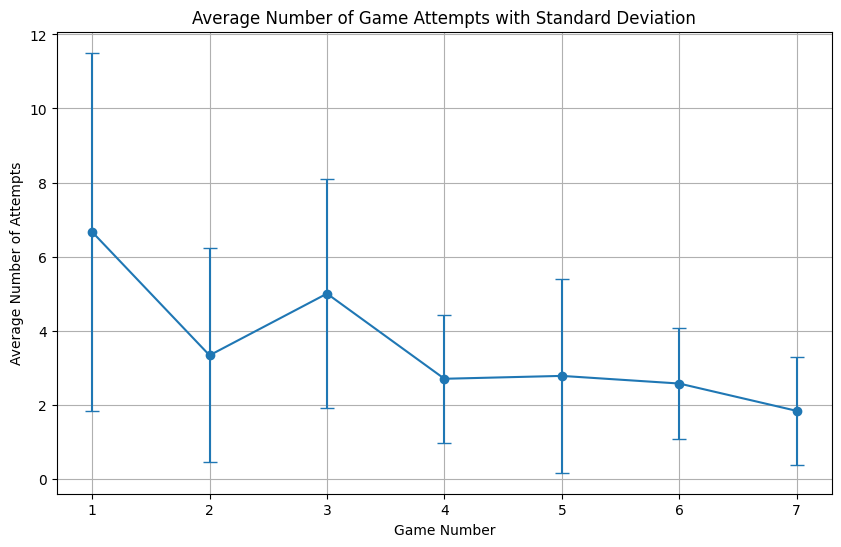} \caption{The graph shows the average number of attempts children took to complete a rhythm-matching game using the AffectaPocket robot in the main study. The error bars represent the standard deviation, indicating the variability in the number of attempts. The descending trend in the average attempts and the narrowing of the error bars over successive games suggest rapid adaptation to tactile cues and a convergence in skill levels across the cohort.}
\label{main_study}
\end{figure}

The main study's results, shown in Figure \ref{main_study}, confirmed our expectations regarding the learning progression. The data obtained from the 18 participants show a trend consistent with our predictions. 

\subsection{On the number of Game Attempts}

The average number of attempts required to complete the rhythm-matching game displayed a descending trend from the first to the seventh game. Initially, the average number of attempts was high, reflecting the children's learning process as they were adapting to tactile feedback without visual cues. As anticipated, there was a significant decrease in the average attempts required as the children played more games, showcasing their quick adaptation and skill acquisition.

\subsection{On the variability in Performance}

In the initial games, there was a wide variance in the number of attempts among the children, documented by the large standard deviation. This variability is indicative of the different initial skill levels, with some children quickly grasping the game mechanics while others taking longer. However, as predicted, the standard deviation decreased with each subsequent game, suggesting a convergence in skill levels as the children became more familiar with the game.

\section{Discussion}

The touch sensors registered two false
positives within a second in one of the four play sessions. When asked
if the participants grasped it during the play session, one
of the children mentioned grasping the device by mistake. This is consistent with the
timing of the recorded touch and release event. There were no further
touch events throughout the first experiment, suggesting that the
capacitive sensors provided a robust way to register when the device is
grasped throughout the day. The instructed grasp events were
registered each half hour even when the robot was concealed in the
pocket of the participants. This could be an argument that the five
metal plates registering each touch event were evenly distributed on the
robot allowing at least one touch plate to be activated with each grasp.
In the full-day 8-hours experiments, 16 touch events were registered at
the beginning of the experiment while the robot was being placed in the
pocket of the participant. No further touch events were registered
throughout the full 8 hours of the experiment showing that the sensors
prove reliable over long-term use. This also indicates that the
sensors worked robustly across a large variation in the activities
performed by the children.

\subsection{Tutorial efficacy}

In the pilot study, the aim was to investigate the efficacy
of the tutorial process. The goal was for the tutorial to aid the children's
transition from using the visual cues on the robot into performing
rhythm-matching with the device concealed in their pockets. The result
show that the skill levels of the children went through similar
progress with each phase and with each new challenge they faced as measured by the number of attempts it took
for them to successfully match each generated rhythm. Through each
phase, every child demonstrated a decreasing number of attempts
to successfully match the notes. Given that the levels of game challenge increased in each phase, seeing 
similar progress and precision for each suggests that the
children quickly adapted to the presented challenges. They progressed
from relying on visual cues to successfully matching the rhythm without
having visual cues, supporting the efficacy of the tutorial process. By following the process, the
children developed a mental model of the game's pattern matching
flow. Importantly, all children learned within a few
attempts, requiring no previous experience or lengthy practice
sessions to be able to use the device. This may be important as anxiety
management techniques often require significant practice to be effective
\autocite{avants1990}. Previous research has shown that although
training often is a beneficial addition to psychotherapy, only 40\% of
those told to practice these measures perform the training as intended.
A possible explanation as to why could be that such repetitive practice
may be uninspiring \autocite{avants1990}. Future
studies could investigate if the gamification aspects of AffectaPocket
provide a stronger motivation for using it. The current threshold for successfully matching each note is
statically set to 40\%. To remain entertaining and challenging, this
threshold should possibly be made more adaptive to each child's skill
level. This would mean that as the child's skill level increases, the
rhythm matches would need to be more precise. Future iterations of the device
could include this and other game modes to potentially reengage the children's interest in using
the robot.

\subsection{Learning progress}

Overall, the decreased number of average attempts and the decreasing standard deviation found in the main study support the efficacy of the visual cues in AffectaPocket robot's tactile game. The similarity of improvements across performance levels and participants suggests that the robot is intuitive for children with varying learning paces. This consistency is critical in a potential therapeutic setting, where the device must accommodate and adapt to each child's unique learning rate without exposing them to many negative experiences.

Analogously, the overall decreasing standard deviation also illustrates that children were able to learn the game and achieve higher skill levels regardless of their individual differences. This is encouraging as it indicates that the robot has the potential to be an effective therapeutic aid for children, regardless of their initial skill levels.

The findings validate the design decision to shorten the tutorial process in the main study, as the children were able to adapt quickly to the tactile cues without extensive practice with visual cues. Such adaptation is critical for therapeutic devices that must be both effective and efficient in their interactions with users. 
\section{Conclusion}

This paper investigated the requirements and feasibility of developing a
pocket robot, an assistive device designed to distract children suffering from an
anxiety disorder by using a rhythm game. A 
pilot study was conducted with typically developing children to examine the robot's main affordances - touch and
tactile features - as well as the practical repercussions of using it the pocket robot a
real-world home setting. The study demonstrated that the robot's touch sensors
reliably detected when a child grasped it, with few false positives,
during different play activities. The study also indicated that
children made excellent progress in comprehending the tactile
capabilities of the robot and were able to use the robot in a concealed
state within minutes. The
robot could be comfortably be carried
throughout an 8-hour study. 
Although the robot has not been tested with a target population, such as children who suffer from anxiety, the functionality of the device has been established through testing with typically developing children. The next step is to conduct a proper evaluation with the target population. 
This work suggests that
AffectaPocket represents a step toward developing devices that can
assist children who experience anxiety and could be complementary to
existing therapy methods. The robot's form provides motivation for children to
use it and may prove to be a readily available, private, and safe means of
redirecting attention during an anxiety attack. 

\textit{Acknowledgements: This research was made possible by the support of the Independent Research Fund Denmark (DFF).}

\balance

\bibliography{root}
\bibliographystyle{IEEEtran}

\end{document}